\documentclass[sigconf]{acmart}

\AtBeginDocument{%
  \providecommand\BibTeX{{%
    \normalfont B\kern-0.5em{\scshape i\kern-0.25em b}\kern-0.8em\TeX}}}

\copyrightyear{2022}
\acmYear{2022}
\setcopyright{rightsretained}
\acmConference[KDD '22]{Proceedings of the 28th ACM SIGKDD Conference on Knowledge Discovery and Data Mining}{August 14--18, 2022}{Washington, DC, USA}
\acmBooktitle{Proceedings of the 28th ACM SIGKDD Conference on Knowledge Discovery and Data Mining (KDD '22), August 14--18, 2022, Washington, DC, USA}
\acmDOI{10.1145/3534678.3539294}
\acmISBN{978-1-4503-9385-0/22/08}

\begin{document}

\title{Semantic Enhanced Text-to-SQL Parsing via Iteratively Learning Schema Linking Graph}





\author{Aiwei Liu}
\affiliation{%
  \institution{Tsinghua University}
  \state{Beijing}
  \country{China}
}
\email{liuaw20@mails.tsinghua.edu.cn}

\author{Xuming Hu}
\affiliation{%
  \institution{Tsinghua University}
  \state{Beijing}
  \country{China}
}
\email{hxm19@mails.tsinghua.edu.cn}

\author{Li Lin}
\affiliation{%
  \institution{Tsinghua University}
  \state{Beijing}
  \country{China}
}
\email{lin-l16@mails.tsinghua.edu.cn}

\author{Lijie Wen}
\authornote{Corresponding author}
\affiliation{%
  \institution{Tsinghua University}
  \state{Beijing}
  \country{China}
}
\email{wenlj@tsinghua.edu.cn}








\renewcommand{\shortauthors}{Aiwei Liu, et al.}
\newcommand{\xuming}[1]{\textcolor{orange}{[#1]}}
\newcommand{\modelname}{\texttt{ISESL-SQL}}
\begin{abstract}

The generalizability to new databases is of vital importance to Text-to-SQL systems which aim to parse human utterances into SQL statements. Existing works achieve this goal by leveraging the exact matching method to identify the lexical matching between the question words and the schema items.  However, these methods fail in other challenging scenarios, such as the synonym substitution in which the surface form differs between the corresponding question words and schema items. 
In this paper, we propose a framework named {\modelname} to iteratively build a semantic enhanced schema-linking graph between question tokens and database schemas.
First, we extract a schema linking graph from PLMs through a probing procedure in an unsupervised manner.  Then the schema linking graph is further optimized during the training process through a deep graph learning method.
Meanwhile, we also design an auxiliary task called graph regularization to improve the schema information mentioned in the schema-linking graph. 
Extensive experiments on three benchmarks demonstrate that {\modelname} could consistently outperform the baselines and further investigations show its generalizability and robustness. 
\end{abstract}




\maketitle

\section{Introduction}
\label{sec:intro}

\begin{figure}
  \includegraphics[width=0.47\textwidth]{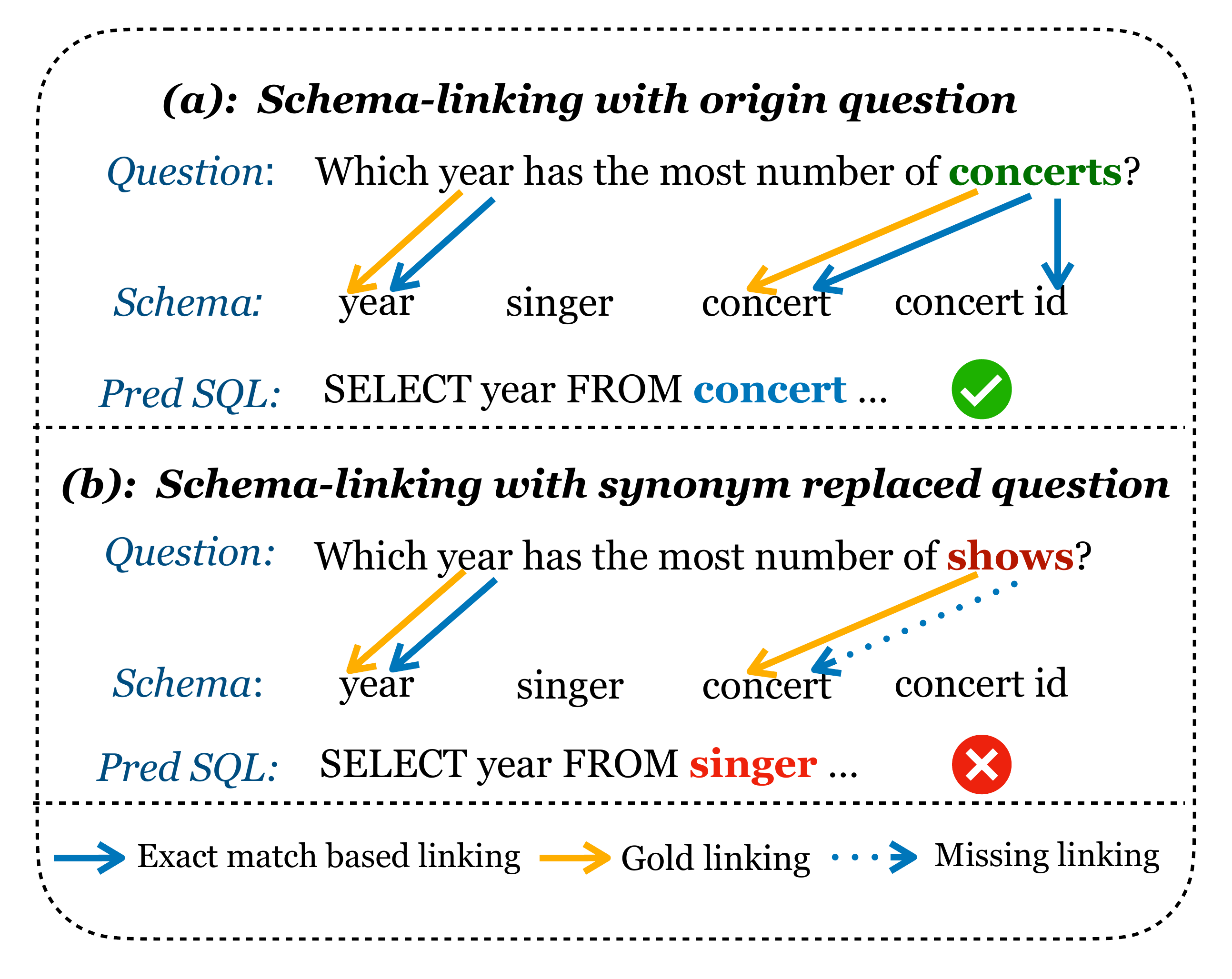}
  \caption{Examples of exact match based schema linking (EMSL). (a): EMSL works fine in the  original question; (b): EMSL fails to identify table \textit{concert} in the question with synonym substitution. }
  \Description{Enjoying the baseball game from the third-base
  seats. Ichiro Suzuki preparing to bat.}
  \label{fig:ESML}
  \vspace{-0.1in}
\end{figure}

As a vast amount of real-world information is stored in databases, using natural language (e.g., English) to inquire information from tabular data is of great importance for modern retrieval applications (e.g., web search engine).
By translating natural language to executable SQL statements, Text-to-SQL becomes an effective solution to achieve this goal \cite{10.1145/3448016.3457543}. Due to databases in the real world being diverse and heterogeneous, the ability for Text-to-SQL systems to be adaptive to different databases instead of being designed for a specific database is of vital importance.  In the single domain setting, the train and test data share the same database, which makes it easy for Text-to-SQL models to learn the relationship between questions and schemas (columns and tables). In contrast to that, databases in train and test data are separated under the cross domain setting, which is challenging for the Text-to-SQL models to infer the schema information mentioned in the question.

Recently, numerous cross-domain Text-to-SQL methods have been proposed, such as IRNet \cite{guo-etal-2019-towards}, RAT-SQL \cite{wang-etal-2020-rat}, LGESQL \cite{cao-etal-2021-lgesql}.
To better extract question-related schema information from databases, all these models use schema linking as the key module to link the tokens in the question to the correct mentioned schema items and thus construct the schema linking graph.  As illustrated in Figure \ref{fig:ESML} (a), a linking is established between question token \textit{concert} and schema item \textit{concerts} to help the Text-to-SQL model find the correct table in the predicted SQL statement. 

However, these methods may fail in real-world settings where schema items are expressed in their synonym form or domain knowledge is required to obtain schema items. As shown in Figure \ref{fig:ESML} (b), when the mention \textit{concerts} is expressed in its synonym form \textit{shows}, the schema linking module fails to detect the correct table \textit{concert},
which leads to the wrong generated SQL statement.  To avoid the limitation of the exact matching based schema linking method, several works explored other methods. 
\citet{dong-etal-2019-data} leverage the implicit supervision from SQL queries to guide the schema linking training by reinforcement learning, which ignores the semantic information and may still fail in synonym substitution settings.  To better capture the semantic relationship between question and schema,  \citet{liu-etal-2021-awakening} train the schema linking module under the supervision of pseudo alignment between token and schema items from PLMs.  However, the supervision from PLMs could be noisy or incomplete.


To obtain more precise and complete semantic linking between questions and schema items, we propose a novel framework named {\modelname} to \textbf{i}teratively build a \textbf{s}emantic \textbf{e}nhanced \textbf{s}chema-\textbf{l}inking graph from PLMs during the Text-to-\textbf{SQL} training process.  We formulate the schema linking procedure as graph construction in which schema linking edges are established between question nodes and schema nodes. Our {\modelname} framework introduces three different modules to construct schema linking graph: initial graph probing module, implicit graph learning module and graph regularization module.
To be robust in challenging settings such as synonym substitution, the initial graph probing module constructs the initial semantic schema linking graph by
masking one question token per time and observing how the PLM embeddings of the schema items are affected.

Because the initial graph is not optimal for the downstream Text-to-SQL task, learning an adaptive graph structure during model training is also needed.  Therefore, the implicit graph learning module learns to generate a sparse implicit weighted graph during training and the whole process is optimized through the downstream Text-to-SQL task in each iteration, which could be viewed as iterative refinement. The implicit weighted graph is then combined with the initial schema linking graph to obtain the schema linking graph in each iteration. We further apply a modified version of  relational graph attention network (RGAT) on it to learn a joint embedding of question and schema nodes.

Although the combined schema linking graph could be optimized by the downstream Text-to-SQL task, whether the model learns the standard schema linking graph is not guaranteed.  Despite the minimal impact on the prediction result, some redundant linking generated by {\modelname} may be difficult to understand, such as the linking between token \textit{concerts} and column \textit{concert id} of the exact match based linking result in Figure \ref{fig:ESML}(a).
To alleviate this problem, the graph regularization module leverages the schema mention information from SQL as implicit supervision to help regulate the schema information.
To summarize, the main contributions of this work are as follows:
\begin{itemize}
    \item We propose a probing method to extract schema linking graphs from PLMs in an unsupervised manner, which makes the Text-to-SQL model more robust under challenging settings like synonym substitution.
     \item To the best of our knowledge, we are the first to introduce the graph structure learning methods into schema linking and Text-to-SQL, which refines the initial schema linking graph iteratively during model training.
    \item We design a graph regularization loss to optimize the schema information in the schema-linking graph.
    \item We show that  {\modelname}  outperforms strong baselines across three benchmarks (Spider, Spider-SYN and Spider-DK).  \footnote{Code and data are available at  \href{https://github.com/THU-BPM/ISESL-SQL}{https://github.com/THU-BPM/ISESL-SQL} } 
\end{itemize}
\begin{figure*}
  \includegraphics[width=0.95\textwidth]{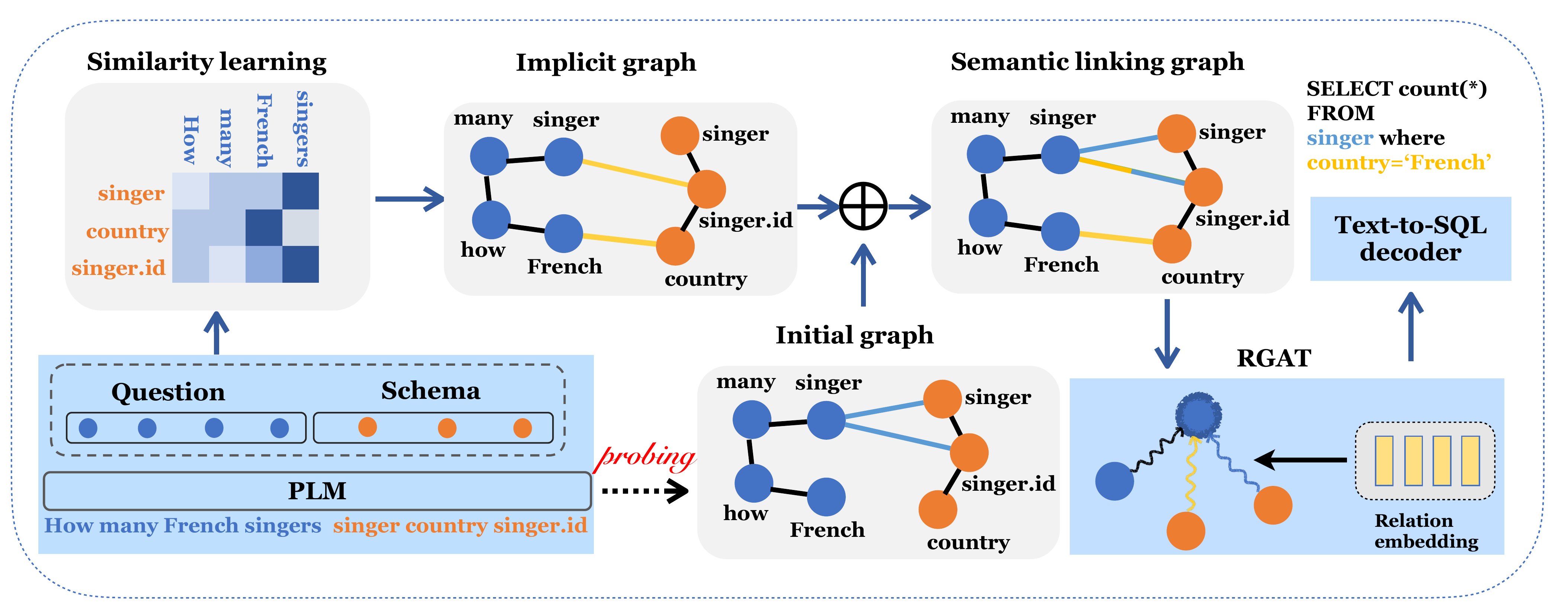}
  \caption{The overall model architecture. We first utilize an unsupervised probing method to construct initial schema-linking graph from PLM (black dashed line). During model training process (blue line), we iteratively learn an adaptive graph structure by parameterized similarity learning. We finally combine these two kinds of graphs and put it into a RGAT network.}
  \Description{Enjoying the baseball game from the third-base
  seats. Ichiro Suzuki preparing to bat.}
  \label{fig:model}
  \vspace{-0.1in}
\end{figure*}

\section{Methods}

The main framework of our proposed {\modelname} is illustrated in Figure \ref{fig:model}, we first probe the initial schema linking graph from PLMs (Section \ref{sec:initial}). Then during the training process, the schema linking graph is iteratively refined (Section \ref{sec:implicit}). A RAGT graph encoder is adopted to encode the schema linking graph(Section \ref{sec:encoder}) and the final SQL statement is generated by a Text-to-SQL Decoder module (Section \ref{sec:decoder}).  A graph regularization module is also applied in the training process (Section \ref{sec:regularization}).

\subsection{Problem Definition}

We first formulate the Text-to-SQL task as follows. Given a natural language question $Q=(q_{1}, q_{2}, \cdots, q_{|Q|})$ and database schema $S$, which contains multiple tables $T=\left\{t_{1}, t_{2},\cdots, t_{|T|}\right\}$ and multi columns $C=\left\{c_{1}, \ldots, c_{|C|}\right\}$, the goal of Text-to-SQL system is to generate a SQL query $y$.  


As a graph encoder is included in our framework, we give a definition to the question-schema graph.  The entire graph can be represented as $G=\left(\mathrm{V}, E\right)$, which contains all three types of mentioned nodes above.  The node sets $V=Q \cup T \cup C$ , in which the number of nodes $\left|V\right|=|Q|+|T|+|C|$. Here, we denote $S= T \cup C$ to represent all the schema nodes. The edge set $E =E_{Q} \cup E_{S} \cup E_{Q \leftrightarrow S} $  contains three kinds of edges: $E_{Q}$, $E_{S}$, and $E_{Q \leftrightarrow S}$. $E_{Q}$ contains the edges inside questions tokens, which are defined by the sequence adjacency relationship, $E_{S}$ includes the edges inside schema items, which are defined by the pre-existing database relations (e.g., primary key relation). Meanwhile, $E_{Q \leftrightarrow S}$ contains the edges between question nodes and schema nodes, which are generated by the schema linking component in Text-to-SQL models. Our work mainly focuses on the construction of $E_{Q \leftrightarrow S}$. 
The total relation matrix of the graph can be defined as follows:

\begin{equation}
\mathbf{E}=\left[\begin{array}{cc}
\mathbf{E}_{Q}^{|Q| \times |Q|} & \mathbf{E}_{Q \leftrightarrow S} \\
\mathbf{E}_{Q \leftrightarrow S}^{T} & \mathbf{E}_{S}^{|S| \times |S|}
\end{array}\right],
\label{eq: def}
\end{equation}
where $|S| = |C| + |T|$ is the length of the schema items. Note that the graph is heterogeneous and the value in $E$ represents the edge type (e.g., 1 for semantic linking relation, 2 for primary key relation).

\subsection{Initial Graph Probing}
\label{sec:initial}

 
The initial Graph Probing module aims to obtain a semantic schema linking graph, which is achieved by masking one question token per time and observing how
the PLM (e.g., BERT \cite{bert-naacl})  embeddings of the schema items are affected. Different from the exact matching based method, the generated graph is not dependent on the surface form of the question tokens.  Specifically, in this section, we aim to generate an adjacency matrix for an edge type $r^{sem}$ in heterogeneous graph  $E_{Q \leftrightarrow S}$ (Eq. \ref{eq: def}).

We first concatenate the question tokens and schema items as $X$:
\begin{align}
X=\left( q_{1}, \cdots, q_{|Q|}, s_{1}, \cdots, s_{|T|+|C|}\right).
\end{align}
The PLM maps the input $X$ into hidden representations. The question and schema embeddings can be denoted as $(\mathbf{q}_{1}, \mathbf{q}_{2}, \ldots, \mathbf{q}_{|Q|})$ and $(\mathbf{s}_{1}, \mathbf{s}_{2}, \ldots, \mathbf{s}_{|T|+|C|})$ respectively.  

Our goal is to induce a function $f\left(q_{i}, s_{j}\right)$ to capture the impact a question word $q_{i}$ has on the representation of schema item $s_{j}$. Inspired by the previous probing method \cite{wu2020perturbed}, we utilize a two-stage approach to achieve this goal in an unsurpervised manner. First, we replace a question token $q_i$ with spacial token [MASK] and feed the new sequence $X \backslash\left\{q_{i}\right\}$ into the PLM again. Then, the representation of schema item $s_{j}$ is changed from $\mathbf{s}_{j}$ to $\mathbf{s}_{j \backslash q_{i}}$, which is affected by the [MASK] of $q_{i}$. We define the $f\left(q_{i}, s_{j}\right)$ by measuring the distance between $\mathbf{s}_{j}$ and $\mathbf{s}_{j \backslash q_{i}}$ as follows:
\begin{align}
f\left(q_{i}, s_{j}\right)=d\left(\mathbf{s}_{j \backslash q_{i}}, \mathbf{s}_{j}\right),
\end{align}
where $d(\mathbf{x}, \mathbf{y})$ is the Euclidean distance between two vectors. 

To obtain a sparse initial schema linking matrix and avoid the noisy edges with low confidence level, we introduce a pre-defined threshold $\tau$ to filter the probing result by:
\begin{equation}
\mathbf{A}^{init}_{i j}=\left\{\begin{array}{ll}
0 & \text { if } f\left(q_{i}, s_{j}\right)  < \tau \\
f\left(q_{i}, s_{j}\right) & \text { if } f\left(q_{i}, s_{j}\right) \ge \tau
\end{array}\right.,
\end{equation}
where $\mathbf{A}^{(init)} \in \mathbb{R}^{|Q| \times (|T|+|C|)}$.


\subsection{Implicit Graph Learning}
\label{sec:implicit}

To repair the noisy and incomplete initial schema linking graph obtained before training, the Implicit Graph Learning module aims to learn an adaptive graph structure and iteratively refine it during model training.

The goal of the implicit graph learning module is to learn an implicit matrix $\mathbf{A}^{(t)} \in \mathbb{R}^{|Q| \times (|T|+|C|)}$ between question tokens and schema items, where $t$ is the epoch number in training. Specifically,  we first calculate the similarity matrix $\mathbf{A}^{(t)} $ by a cosine similarity based function as follows:
\begin{equation}
\mathbf{A}^{(t)}_{i j}=ReLU\left(\frac{<\mathbf{q}_{i} \times \mathbf{W}_{1}, \mathbf{s}_{j} \times \mathbf{W}_{2}>}{| \mathbf{q}_{i} \times \mathbf{W}_{1} | |\mathbf{s}_{j} \times \mathbf{W}_{2}|}\right),
\label{eq: sim}
\end{equation}
where $ReLU(x)$ is the rectified linear unit activation function \cite{glorot2011deep} adopted to retain the positive part of its argument. 
$\mathbf{W}_{1}$ and $\mathbf{W}_{2}$ are two learnable weight vectors that transform the question/schema embeddings for similarity calculation and $<\mathbf{x},\mathbf{y}>$ denotes vector inner product operation.

We could get an weighted similarity matrix from the function in Eq.\ref{eq: sim}. In practice, most schema items only link to one question node,  so we propose a graph sparsification component which only keeps the max similarity score for each schema. 
Our graph sparsification component is defined as follows:
\begin{equation}
\mathbf{A}^{t}_{i j}=\left\{\begin{aligned}
\mathbf{A}^{t}_{i j} , \quad & \mathbf{A}^{t}_{i j}= \operatorname{Max}\left(\mathbf{A}^{t}_{j}\right) \\
0,\quad &  \mathbf{A}^{t}_{i j} \neq \operatorname{Max}\left(\mathbf{A}^{t}_{j}\right)
\end{aligned}\right.,
\label{eq:implict}
\end{equation}
where $A^{t}_{j} =\left[A^{t}_{0,j}, \ldots, A^{t}_{|Q-1|,j}\right]$ is the similarity vector of each schema based in Eq.\ref{eq: sim}.


\subsection{Graph Encoder}
\label{sec:encoder}
Given the initial schema semantic linking matrix $\mathbf{A}^{init}$ and the implicit semantic schema-linking matrix  $\mathbf{A}^{(t)}$, the graph encoder module tries to learn a better question token and schema item embedding with the enhanced schema linking graph, which could help Text-to-SQL decoder to identify the correct schema item. 

The schema linking graph construction procedure at epoch $t$ can be represented as:
\begin{equation}
\widetilde{\mathbf{A}}^{(t)}=\lambda \mathbf{A}^{init}+(1-\lambda) \mathbf{A}^{(t)}
\label{eq: combine}
\end{equation}
where $\lambda$ is a hyperparameter used to adjust the weights of the two graphs.  As shown in Eq.\ref{eq: def}, the input graph of the graph network is a heterogeneous network which contains different edge types (e.g., semantic linking relation between the question and schema and primary key relation inside schema items).      $\widetilde{\mathbf{A}}^{(t)}$ can be seen as a weighed matrix of a specific relation $r^{sem}$. Given $\widetilde{\mathbf{A}}^{(t)}$, we update the graph $E_{Q \leftrightarrow S} $ in Eq.\ref{eq: def} as:
\begin{equation}
[\mathbf{E}_{Q \leftrightarrow S}]_{ij}^{(t)}=\left\{\begin{aligned}
r^{sem},  \quad& \widetilde{\mathbf{A}}^{(t)} > 0 \\
r^{none}, \quad&  \widetilde{\mathbf{A}}^{(t)} = 0
\end{aligned}\right.
\end{equation}
In epoch t, the whole graph  $\mathbf{E}^{(t)}$ is then generated by replacing $\mathbf{E}_{Q \leftrightarrow S}$ in Eq.\ref{eq: def} by  $\mathbf{E}_{Q \leftrightarrow S}^{(t)}$.
Then we expand the weighted matrix $\widetilde{\mathbf{A}}^{(t)}$ to the whole graph, which can be represented as:
\begin{equation}
\mathbf{M^{(t)}}=\left[\begin{array}{cc}
\mathrm{1}^{|Q| \times |Q|} & \widetilde{\mathbf{A}}^{(t)} \\
\widetilde{\mathbf{A}}^{(t)^{T}} & \mathrm{1}^{|S| \times |S|}
\end{array}\right]
\label{eq: def1}
\end{equation}
As shown in Eq.\ref{eq: def1}, we only calculate the weight matrix between question and schema. The weight of other graphs can be directly set to 1 because there is no uncertainty in these edges.

To process the input heterogeneous graph $E$, we introduce a relational graph attention network (RGAT) \cite{wang-etal-2020-relational} as our graph encoder.  The RGAT network utilizes a learnable relational embedding to control the different impacts among edge types.  To simplify, we use $\mathbf{X}^{l} \in \mathbb{R}^{(\left|Q\right|+\left|C\right|+\left|T\right|) \times d}$,  to denote the entire node embedding matrix in layer $l$ where d is the GNN embedding size. Similar to previous works \cite{wang-etal-2020-rat, cao-etal-2021-lgesql}, we utilize the multi-head scaled dot-product for attention weights calculation.  To optimize the  weighted matrix $M^{(t)}$ in each epoch,  different from previous works, we introduce $M^{(t)}$ to the graph attention calculation process. 

Specifically, given the current node representations $\mathbf{X}^{l}$, graph $E^{(t)}$ and weighted matrix $M^{(t)}$, 
the attention weight $\alpha_{j i}^{h}$ is calculated by following equations:
\begin{equation}
\alpha_{j i}=\left(\mathbf{x}_{i} \mathbf{W}_{q}\right)\left(\mathbf{x}_{j} \mathbf{W}_{k}+\mathbf{M}_{ji} \left[ F \left(E_{j i}\right)\right]\right)^{\mathrm{T}},
\label{eq:attention1}
\end{equation}

\begin{equation}
\alpha_{j i}^{h}=\operatorname{softmax}_{j}\left(\hat{\alpha}_{j i}^{h} / \sqrt{d}\right),
\end{equation}
where matrices $\mathbf{W}_{q}^{h}, \mathbf{W}_{k}^{h}, \mathbf{W}_{v}^{h} \in \mathbb{R}^{d \times d / H}$ are trainable transformations and H is the number of heads. Function $\psi\left(\right)$ is a learnable matrix which transforms the relation type $E_{j i}^{(t)}$ into a d-dim feature vector.  $\mathbf{M}^{(t)}_{ji}$ is used to control the influence of relation embedding by a simple multiplication operation. Operator $[\cdot]_{h}^{H}$ first evenly splits the vector into $H$ parts and returns the $h$-th partition.  

Then the output representation of current layer $\mathbf{x}_{i}^{l+1}$ is generated by calculating the relation embedding in a similar way:

\begin{equation}
\mathbf{x}_{i} =\sum \alpha_{j i}\left(\mathbf{x}_{j} \mathbf{W}_{v}+  \mathbf{M}_{ji}\left[F\left(E_{j i}\right)\right]\right)
\label{eq:attention2}
\end{equation}
\begin{equation}
{\mathbf{x}}_{i}^{l+1}=FFN\left(\operatorname{LayerNorm}\left(\mathbf{x}_{i}^{l}+\hat{\mathbf{x}}_{i}^{l} \mathbf{W}_{o}\right)\right)
\label{eq:attention3}
\end{equation}
where matrices  $\mathbf{W}_{o} \in \mathbb{R}^{d \times d}$ is the output transformation, $\|$ represents vector concatenation operation and FFN $(\cdot)$ denotes one feedforward neural network layer. 


Since the weighted matrix $M^{(t)}$ is optimized in each epoch, our schema linking graph could be updated iteratively during model training. 

The final output of the graph encoder $\mathbf{x^{e}}$ can be split into question representation $(\mathbf{q}_{1}^{e}, \mathbf{q}_{2}^{e}, \ldots, \mathbf{q}_{|Q|}^{e})$ and schema representation $(\mathbf{s}_{1}^{e}, \mathbf{s}_{2}^{e}, \ldots, \mathbf{s}_{|T|+|C|}^{e})$ for the computation of Text-to-SQL decoder.

\subsection{Text-to-SQL Decoder}
\label{sec:decoder}

Given the question representation $(\mathbf{q}_{1}^{e}, \mathbf{q}_{2}^{e}, \ldots, \mathbf{q}_{|Q|}^{e})$ and schema representation $(\mathbf{s}_{1}^{e}, \mathbf{s}_{2}^{e}, \ldots, \mathbf{s}_{|T|+|C|}^{e})$ generated by graph encoder module, Text-to-SQL decoder module aims to generate the corresponding SQL statements.

The architecture of our Text-to-SQL decoder is similar to previous work \cite{yin-neubig-2017-syntactic}, which generates the abstract syntax tree (AST) of the target SQL $y$ in depth-first traversal order.  The action generated by decoder in each timestep is either: (1) An APPLYRULE action that expands the current non-terminal node based on SQL grammar in the partially generated AST. (2) A SELECTTABLE or SELECTCOLUMN action that chooses one table or column item from the encoded schema representation $(\mathbf{s}_{1}^{e}, \mathbf{s}_{2}^{e}, \ldots, \mathbf{s}_{|T|+|C|}^{e})$ \footnote{More implementation details are provided in appendix}.

\subsection{Graph Regularization}
\label{sec:regularization}


Although the combined graph of the initial graph $\mathbf{A}^{init}$ and implicit graph $\mathbf{A}^{(t)}$ could approach the optimized graph iteratively, the quality of the implicit graph is not guaranteed. Due to that the schema linking module is to find the mentioned schema in question, it is more important for the implicit graph to find the correct schema items than question tokens. In the following, we introduce the auxiliary graph regularization loss to further optimize the schema linking graph

For each SQL query $y$, we denote the set of column and table names appearing in $y$ as $S_{SQL}$, which consists of columns and table mentions, i.e., $S_{SQL}$ = $S_{COL} \cup S_{TBL}$. During the training process, we can directly utilize $S_{SQL}$ as the supervision of our graph regularization loss.  Specifically,  we make the sum of $\mathbf{A}^{(t)}$ in the question dimension approximate the $S_{SQL}$ by binary cross-entropy loss.
\begin{equation}
   \mathcal{L}_{g} =  - \sum_{j \in S_{SQL}}log(\sum_{i}A^{(t)}_{ij})
   \label{eq:loss}
\end{equation}
Note that in Eq.\ref{eq:implict}, only the largest value item of question dimension in $\mathbf{A}^{(t)}$ is retrained. Therefore, the graph regularization loss actually only optimizes the most relevant question of mentioned schema. 

This graph regularization auxiliary loss is combined with the main Text-to-SQL loss as follows:
\begin{equation}
\mathcal{L} = \mathcal{L}_{\mathrm{SQL}} + \mu \mathcal{L}_{g} 
\label{eq:final}
\end{equation}
where $\mu$ is a hyperparameter that balances the absolute value of the two losses.


\section{Experiments}

In this section, we conduct extensive experiments on three different benchmarks to evaluate the effectiveness of our proposed {\modelname} model and give detailed analyses to show its advantage.
\subsection{Experiment Setup}

\subsubsection{Datasets Description.} We conduct extensive experiments on three public benchmark datasets as follows: (1) \textbf{Spider} \cite{yu-etal-2018-spider} is a large-scale cross-domain Text-to-SQL benchmark. It contains 8659 training samples across 146 databases and 1034 evaluation samples across 20 databases. We report the exact set match accuracy on the development set, as the test set is not publicly available. (2) \textbf{Spider-DK} \cite{gan-etal-2021-exploring} is a human-curated dataset based on Spider, which samples 535 question-SQL pairs across 10 databases from Spider development set and modifies them to incorporate the
domain knowledge. The schema information is implicitly expressed in Spider-DK and thus complex reasoning is required. We train our model on spider training set and test on the Spider-DK development set. (3) \textbf{Spider-SYN} \cite{gan-etal-2021-towards} is another challenging variant of the Spider dataset.  Spider-SYN is constructed by manually modifying natural language questions with synonym substitution, which is more adaptable for the scenario where users do not know the exact schema words mentioned. There are total 1034 samples across 20 databases in Spider-SYN.  Our model is trained on the  spider training dataset and tested on the Spider-SYN development set.

\subsubsection{Baselines} We compare our proposed model with several competing baselines. (1) \textbf{LGESQL} \cite{cao-etal-2021-lgesql} is a graph attention network based sequence-to-sequence model with relational GAT and the line graph, which is the previous state-of-the-art Text-to-SQL model. (2) \textbf{RATSQL} \cite{wang-etal-2020-rat} is a sequence-to-sequence model enhanced by a relational-aware transformer. (3) \textbf{ETA} \cite{liu-etal-2021-awakening} also aims to fix the exact match based schema linking with the pseudo labels generated by PLMs.  (4) \textbf{SmBoP} \cite{rubin-berant-2021-smbop} introduces a semi-autoregressive bottom-up decoder to generate SQL statements. (5)  \textbf{DT-Fixup SQL-SP} \cite{DBLP:conf/acl/Xu0YZT0CPC20}  proposes a theoretically justified optimization strategy to train Text-to-SQL model. (6) \textbf{IRNET} \cite{guo-etal-2019-towards} proposes an intermediate representation SemSQL to close the gap between natural language and SQL statements. (7) \textbf{EditSQL}  \cite{zhang2019editing} views SQL as sequences and reuses previous generation results at the token level.  (8) \textbf{RYANSQL} \cite{DBLP:journals/coling/ChoiSKS21} generates nested queries by recursively yielding its component \texttt{SELECT} statements and uses a sketch-based slot filling approach to predict each \texttt{SELECT} statement. (9) \textbf{Global-GNN} \cite{bogin-etal-2019-global}  proposes a semantic parser that globally reasons about the structure of the output query to make a more contextually informed selection of database constants. 
Many previous works design adaptive PLM models for specific Text-to-SQL models to achieve better result, such as GAP \cite{DBLP:conf/aaai/ShiNWZLWSX21}, GRAPPA \cite{DBLP:conf/iclr/0009WLWTYRSX21}, STRUG \cite{DBLP:conf/naacl/DengAMPSR21}. For fair comparison, except for comparing the result on a unified pre-training model  {\texttt{BERT-large}}, we also report the result with model adaptive PLM.


\subsubsection{Hyper-parameters} The threshold $\tau$ in the initial graph probing process is set to $0.7$.  In the encoder part, the hidden size $d$ of the RGAT is $256$ for GloVe and $512$ for all PLMs. The number of RAGT layers $L$ is $8$.  The number of heads in RGAT's multi-head attention is $8$ and the dropout rate of features is set to $0.2$. $\lambda$ is set to $0.2$ when combining the initial graph and implicit graph. In the decoder part, following the previous work \cite{cao-etal-2021-lgesql}, the dimension of hidden state, action embedding, and node type embedding size is set to $512$, $128$, and $128$ respectively.  And the dropout rate for decoder LSTM is $0.2$. We use AdamW \cite{DBLP:conf/iclr/LoshchilovH19} with learning rate $5e$-$4$ for GloVe and $1e$-$4$ for PLMs. The balancing hyperparameter $\mu$ is set to 1 in Eq.\ref{eq:final}. Furthermore, we use a linear warmup scheduler with a warmup ratio of $0.1$. The batch size is set to $20$ and the total training epoch is $200$.

\subsection {Main Results}
Table \ref{tab:main-result} shows the exact match accuracy on three benchmarks with the exact match average accuracy of 3 runs. LGESQL is the previous state-of-the-art model in all three embedding configurations (without PLM, with {\texttt{BERT-large}} and with model adaptive PLM). 
In general,  {\modelname} could outperform previous baselines in all configurations. More specifically, with GloVe word vectors, our model could surpass the previous best model by an average of $2.5\%$ over all benchmarks. Similarly, {\modelname} could achieve an average performance boost of $1.3\%$ with a unified pre-training model  {\texttt{BERT-large}} over the three benchmarks. Furthermore, our {\modelname} achieves state-of-the-art performance with {\texttt{ELECTRA-large}} \cite{DBLP:conf/iclr/ClarkLLM20} on the model adaptive PLM setting.

We observe that {\modelname} could obtain greater improvement on Spider-SYN and Spider-DK benchmarks than standard spider benchmark, which proves the robustness of {\modelname} in challenging datasets. Also, since the ability for  GloVe word vectors to capture semantic linking is inferior to PLM embeddings, {\modelname} achieves more performance boost with GloVe word vectors setting.

\begin{table}[bt!]

\resizebox{0.93\linewidth}{!}{
\begin{tabular}{l|c|c|c}
\toprule
\multicolumn{1}{c}{\textbf{Model}} & \multicolumn{1}{c}{\textbf{Spider}} & \multicolumn{1}{c}{\textbf{Spider-DK}} & \multicolumn{1}{c}{\textbf{Spider-SYN}}   \\
\midrule 
\midrule 
\multicolumn{4}{c}{\textbf{Without PLM: GloVe}} \\
\midrule
Global-GNN + GloVe \cite{bogin-etal-2019-global}   &52.7  &26.0 & 23.6 \\
EditSQL + GloVe\cite{zhang2019editing}     &36.4 & 31.4 &25.3 \\
IRNet + GloVe\cite{guo-etal-2019-towards}  &53.2 & 33.1 &28.4  \\
RATSQL + GloVe \cite{wang-etal-2020-rat} &62.7 & 35.8  & 33.6\\
LGESQL + GloVe \cite{cao-etal-2021-lgesql}  &67.6 & 39.2 &40.5  \\
\midrule 
\textbf{ISESL-SQL + GloVe}   &  68.3 (0.7$\uparrow$) &  42.1 (2.9$\uparrow$) &  44.4 (3.9$\uparrow$)   \\
\midrule 
\midrule 
\multicolumn{4}{c}{\textbf{With PLM: BERT}} \\
\midrule
EditSQL + BERT-large \cite{zhang2019editing}      &57.6 &36.2 & 34.6  \\
IRNet + BERT-large\cite{guo-etal-2019-towards}   &53.2  &38.6 & 36.7  \\
RYANSQL + BERT-large \cite{DBLP:journals/coling/ChoiSKS21}  &66.6 &40.1 &47.8 \\
RATSQL + BERT-large \cite{wang-etal-2020-rat}  &69.7 &40.9 & 48.2 \\
ETA + BERT-large \cite{liu-etal-2021-awakening}   &70.8 &41.8 &50.6 \\
LGESQL + BERT-large  \cite{cao-etal-2021-lgesql} &74.1  &44.7 &55.1  \\
\midrule 
\textbf{ISESL-SQL + BERT-large}    &  74.7 (0.6 $\uparrow$) &  46.2 (1.5$\uparrow$) & 56.8 (1.7$\uparrow$)   \\
\midrule 
\midrule 
\multicolumn{4}{c}{\textbf{With Model Adaptive PLM}} \\
\midrule
RATSQL + STRUG  \cite{DBLP:conf/naacl/DengAMPSR21}    &72.6 &39.4 & 48.9  \\
RATSQL + GRAPPA  \cite{DBLP:conf/iclr/0009WLWTYRSX21}  &73.4 & 38.5 &49.1\\
SmBoP + GRAPPA  \cite{rubin-berant-2021-smbop} & 74.7 & 42.2 &48.6  \\
RATSQL + GAP  \cite{DBLP:conf/aaai/ShiNWZLWSX21} & 71.8 & 44.1 &49.8  \\
DT-Fixup SQL-SP + RobERTa \cite{DBLP:conf/acl/Xu0YZT0CPC20}    & 75.0 &40.5 &50.4  \\
LGESQL + ELECTRA-large \cite{cao-etal-2021-lgesql}   &75.1 & 48.4 &60.0  \\
\midrule 
\textbf{ISESL-SQL + ELECTRA-large}    & \textbf{75.8} (0.7$\uparrow$)  &  \textbf{50.0} (1.6$\uparrow$)  & \textbf{62.2} (2.2 $\uparrow$)   \\ 

\bottomrule
\end{tabular}}
\caption{Exact match accuracy (\%) on three different benchmarks: Spider,Spider-DK and Spider-SYN. }
\label{tab:main-result}
\vspace{-0.3in}
\end{table}

\subsection {Ablation Studies}

We conduct ablation studies to show the effectiveness of different modules of {\modelname} to the overall improved performance.  {\modelname} \texttt{w/o initial graph pruning} is the proposed model without the initial probed graph and only keeps the learned graph during training.   {\modelname}  \texttt{w/o implicit graph learning} only adopts the initial probed graph with no more graph optimization process.  {\modelname}  \texttt{w/o schema linking} replaces all the edges in $\mathbf{E}_{Q \leftrightarrow S}$ (Eq.\ref{eq: def}) to none-relation edge type $r^{none}$.  {\modelname}  \texttt{w/o graph regularization} removes the graph regularization loss and only keeps the Text-to-SQL loss during training.  {\modelname}  \texttt{w exact match schema linking} replaces our semantic schema linking edge with the the exact match schema linking edges. 


A general conclusion from ablation results in Table \ref{tab:ablation} is that all modules contribute positively to the improved performance. More specifically, {\modelname} \texttt{w/o initial graph probing}  gives us $1.4\%$ less performance averaged on all datasets. Similarly, implicit graph learning gives $1.3\%$ performance boost in average over all benchmarks. Removing the graph regularization loss leads to an average of $0.9\%$ performance drop. Furthermore, when totally removing the schema linking edges,  {\modelname} \texttt{w/o schema linking} brings an average performance drop of $1.4\%$.  Compared with the exact match schema linking method, our {\modelname} gives an average improvement of $2.4\%$.

We can discover that the  initial graph probing module contributes more in spider-SYN and  spider-DK than  spider benchmark, which proves the importance of semantic linking edges in challenging settings. Also, the exact match schema linking method is even worse than no schema linking settings, which shows it vulnerability in challenging settings.

\begin{table}[bt!]

\resizebox{0.99\linewidth}{!}{
\begin{tabular}{l|c|c|c}
\toprule
\multicolumn{1}{c}{\textbf{Technique}} & \multicolumn{1}{c}{\textbf{Spider}} & \multicolumn{1}{c}{\textbf{Spider-SYN}} & \multicolumn{1}{c}{\textbf{Spider-DK}}  \\
\midrule 
\midrule 
\modelname & \textbf{75.8} & \textbf{62.2} & \textbf{50.0}\\
w/o initial graph probing  & 75.0  (0.8$\downarrow$) & 60.2  (2.0$\downarrow$) & 48.5(1.5$\downarrow$)  \\
w/o implicit graph learning & 74.6 (1.2$\downarrow$)  & 60.7 (1.5$\downarrow$) & 48.7(1.3$\downarrow$) \\
w/o schema linking & 73.4  (2.5$\downarrow$) & 61.2  (1.0$\downarrow$) &  49.3 (0.7$\downarrow$) \\
w/o graph regularization  & 74.8 (1.0$\downarrow$ ) & 61.3  (0.9$\downarrow$) &  49.2 (0.8$\downarrow$) \\
w exact match schema linking  & 73.6  (2.2$\downarrow$)  & 59.7  (2.5$\downarrow$) & 47.6  (2.4$\downarrow$)\\
\bottomrule
\end{tabular}}
\caption{Ablation study of different modules.}
\label{tab:ablation}
\vspace{-0.2in}
\end{table}
\subsection {Schema Linking Analysis}

\begin{table}[bt!]

\resizebox{\linewidth}{!}{
\begin{tabular}{l|c|c|c|c|c|c}
\toprule
\multicolumn{1}{c}{\textbf{Model}} & \multicolumn{1}{c}{\textbf{Col$_P$}} & \multicolumn{1}{c}{\textbf{Col$_R$}} & \multicolumn{1}{c}{\textbf{Col$_F$}} & \multicolumn{1}{c}{\textbf{Tab$_P$}} & \multicolumn{1}{c}{\textbf{Tab$_R$}} & \multicolumn{1}{c}{\textbf{Tab$_F$}}   \\
\midrule 
\midrule 
N-gram Match  &61.4  &69.4 &65.1 &78.2  &69.6 & 73.6 \\
SIM   &16.6 & 8.0  & 10.8 & 8.5  &11.6 & 9.8 \\
CONTRAST  &83.7  &68.4 &75.3 & 84.0 &76.9 &80.3 \\
ETA   &86.1 & 79.3 & 82.5 & 81.1  & 85.3 & 83.1\\
SLSQL$_{L}^{\heartsuit}$   &82.6 &82.0 &82.3 &80.6 &84.0 &82.2 \\
\midrule 
\textbf{ISESL-SQL}& \textbf{87.2}(1.1$\uparrow$) & \textbf{85.3}(3.3$\uparrow$)  & \textbf{86.2}(3.7$\uparrow$)  & 
\textbf{89.4}(5.4$\uparrow$) & \textbf{87.1}(1.8$\uparrow$)  & \textbf{88.2}(5.1$\uparrow$) \\
\bottomrule
\end{tabular}}
\caption{Schema linking experimental results on spider dev sets. $^{\heartsuit}$ means the model uses schema linking supervision.}
\label{tab:schema-linking}
\vspace{-0.2in}
\end{table}

To better demonstrate the quality of our constructed schema linking graph,  we compare the schema information (tables and columns) produced by {\modelname} with human annotations. 

Following the previous works \cite{lei-etal-2020-examining, liu-etal-2021-awakening}, we report the micro average precision, recall and F1-score for both columns ($Col_P$ , $Col_R$, $Col_F$) and tables ($Tab_P$ ,$Tab_R$, $Tab_F$). The metric focus on whether the correct schema item is identified.

We consider five strong baselines for comparison. (1) \textbf{N-gram Matching} links all n-gram ($n \leq 5$) phrases in a natural language question to schema items
by fuzzy string matching. (2) \textbf{SIM} computes the cosine similarity between question tokens and schema items using PLM embeddings without task specific fine-tuning. (3) \textbf{CONTRAST} learns by comparing the aggregated embedding scores of mentioned schemas with unmentioned ones in a contrastive learning style, as done in \citet{liu-etal-2020-impress}. (4) \textbf{SLSQL$_{L}$} \cite{lei-etal-2020-examining} is trained with full schema linking supervision  by a learnable schema linking module. (5) \textbf{ETA} \cite{liu-etal-2021-awakening} trains the schema linking module using pseudo alignment as supervision generated from PLMs.

Table \ref{tab:schema-linking} shows the experimental results on the schema linking accuracy. Our proposed {\modelname} model could surpass all weakly supervised and unsupervised methods by a large margin.  Specifically, our method brings an improvement of $3.7\%$ on $Col_F$  and $5.1\%$ on $Tab_F$ over the previous best baseline ETA \cite{liu-etal-2021-awakening}. SIM is a similar method to our initial graph probing component which constructs schema linking graph in an unsupervised way.  We discover that our method outperforms SIM by a  huge margin, which indicates the effectiveness of our iterative graph learning process. Surprisingly, our  {\modelname} method could outperform SLSQL$_{L}$ which is trained in a fully supervised manner. This indicates that the weak supervision from schema information could achieve a similar result with fully supervision. 

\begin{figure}[t!]
    \centering
    \includegraphics[width=0.49\linewidth]{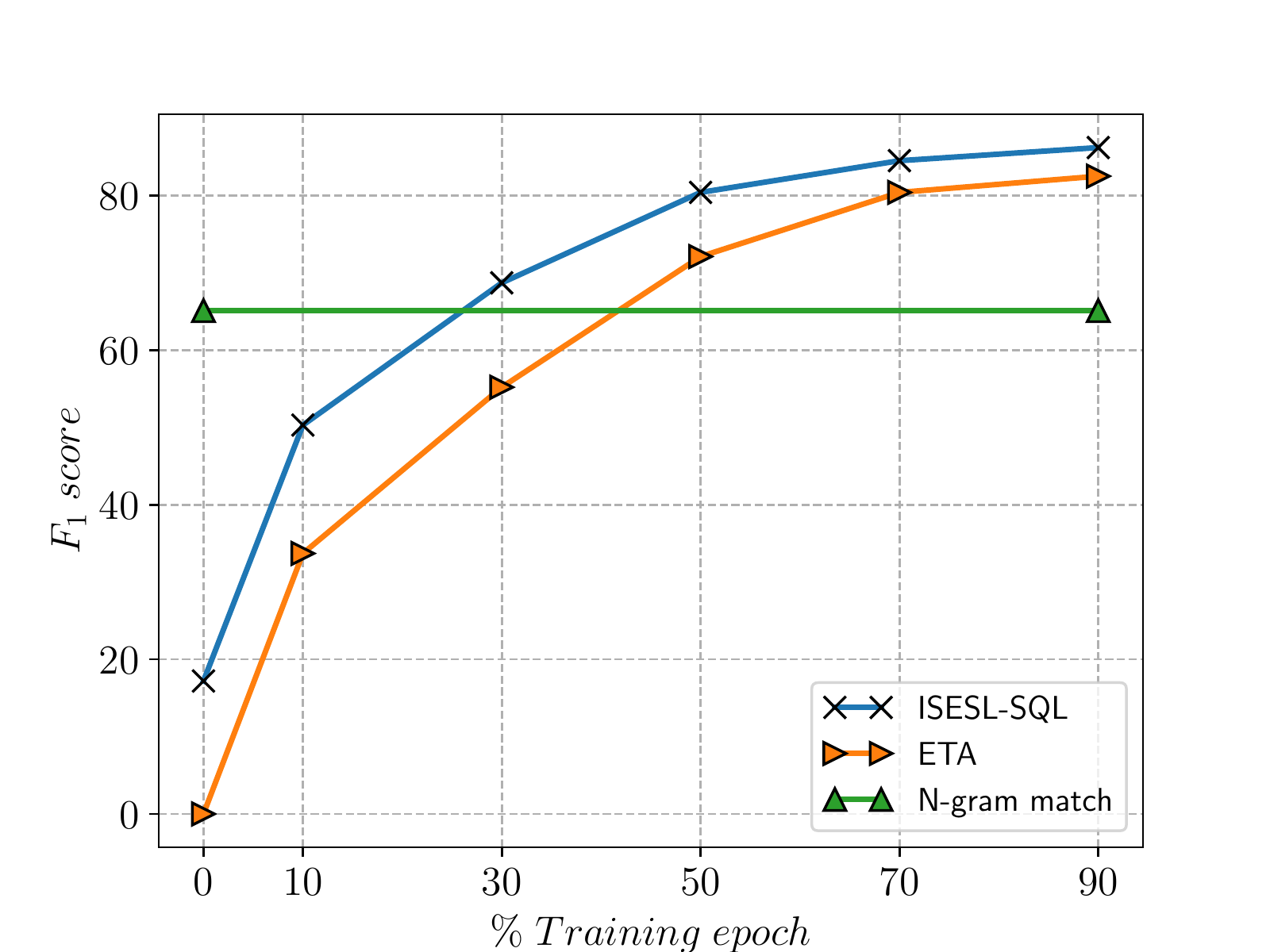}
    \includegraphics[width=0.49\linewidth]{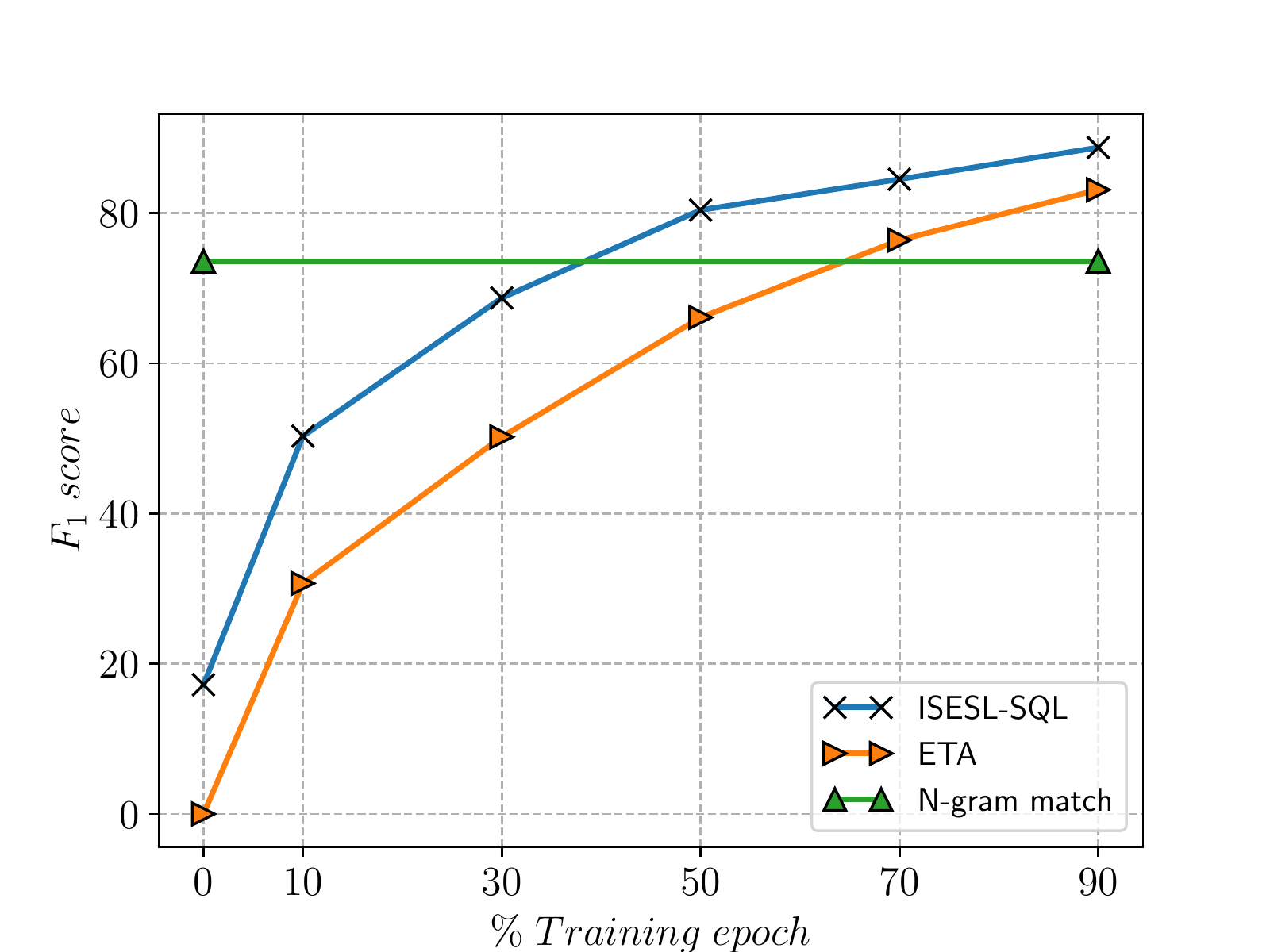}
    \caption{F1 results of the column match (left) and table match (right) accuracy during the training process on spider development set. }
    \label{fig:trend}
\end{figure}

To further investigate how our model learns the schema linking graph iteratively during the training process, we visualize the changing trend of schema match F1 score during the training process in the spider development set. As shown in Figure \ref{fig:trend}, our method could outperform the previous best baseline ETA during the entire training process. And after epoch 45, our {\modelname}  surpasses the N-gram match based method. Our model could achieve the best schema linking result before epoch 100 in both column matching and table matching. Also, it can be seen that our {\modelname} model has an initial F1 score before training, which is the result of the initial graph probing. These results prove our model could iteratively refine the schema linking graph during the training process.

\subsection {Component Matching Analysis}

\begin{table}[bt!]

\resizebox{0.86\linewidth}{!}{
\begin{tabular}{l|c|c|c}
\toprule
\multicolumn{1}{c}{\textbf{SQL component}} & \multicolumn{1}{c}{\textbf{RATSQL}} & \multicolumn{1}{c}{\textbf{LGESQL}} & \multicolumn{1}{c}{\textbf{ISESL-SQL}}  \\
\midrule 
\midrule 
SELECT  & 74.6  & 84.0 & 84.7 (0.7 $\uparrow$) \\
SELECT (no AGG)  & 76.3 & 85.2 & 86.3 (1.1 $\uparrow$)\\
WHERE  & 73.1  & 71.8 & 74.7 (1.6 $\uparrow$ ) \\
WHERE(no OP) & 77.3 & 76.2 & 79.5 (2.2 $\uparrow$)\\
GROUP BY (no HAVING)  & 67.0 & 81.6 & 81.0 (0.6 $\downarrow$) \\
GROUP BY   & 63.1 & 79.4 & 78.3 (1.1 $\downarrow$) \\
ORDER BY   &79.2 & 82.9 & 83.0 (0.1 $\uparrow$ ) \\
AND/OR   &98.3 & 97.7 & 98.1 (0.2 $\downarrow$ )\\
IUE   &27.5 & 51.2 & 50.3 (0.9 $\downarrow$ ) \\
KEYWORDS   &87.4 &87.3 & 88.5 (1.1 $\uparrow$ ) \\
\bottomrule
\end{tabular}}
\caption{F1 scores of component matching of RATSQL, LGESQL and our model on Spider-SYN benchmark.}
\label{tab:detail}
\vspace{-0.3in}
\end{table}

To better analyze the reasons for the performance improvements achieved by our {\modelname}, we perform a detailed analysis of the matching accuracy of different components of the SQL statements on the Spider-SYN benchmark. As shown in Table \ref{tab:detail}, the performance improvement mainly comes from the components including schema items, such as \textit{SELECT}, \textit{SELECT (no AGG)}, \textit{WHERE} and \textit{WHERE (op)}.  Specifically,  on the \textit{WHERE (op)} component,  our  {\modelname}   model outperforms the previous best baseline by $2.2\%$. On other components where schema items are not included,  the performance remains unchanged or slightly drops on some components, such as \textit{IEU} and \textit{AND/OR}. The observation shows a high-quality schema linking graph could help the Text-to-SQL model find the correct schema items.

\begin{figure*}
  \includegraphics[width=0.89 \textwidth]{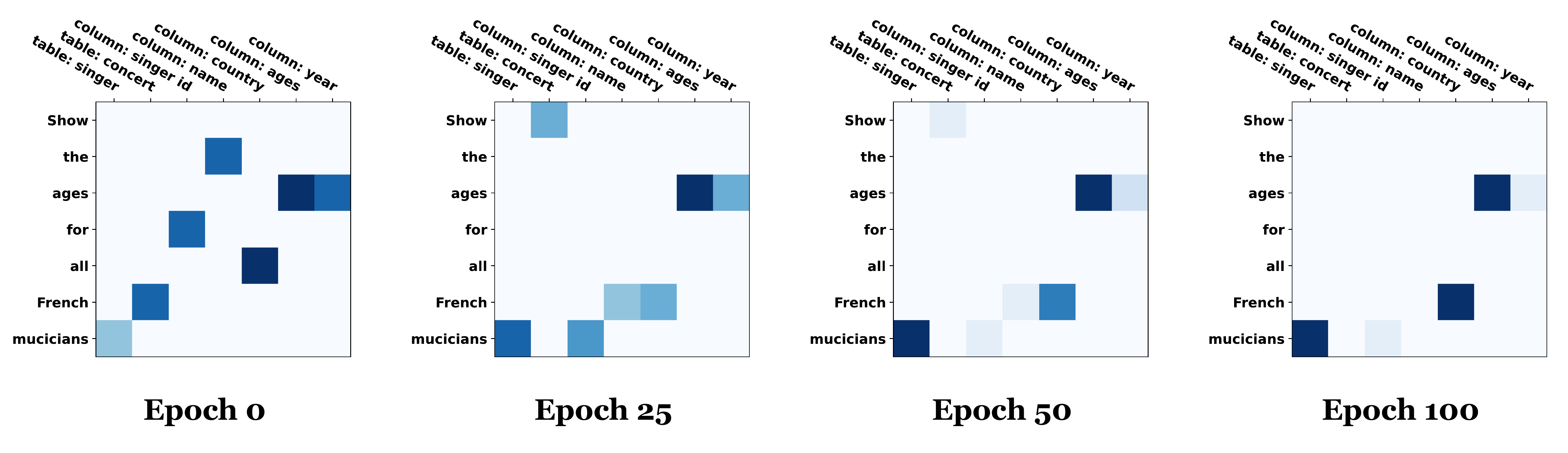}
  \vspace{-0.1in}
  \caption{Alignment matrix between the question ``show the ages for all French musicians'' and the database `concert\_singer' schema in training epoch 0, 25, 50, and 100.}
  \Description{Enjoying the baseball game from the third-base
  seats. Ichiro Suzuki preparing to bat.}
  \label{fig:case}
  \vspace{-0.1in}
\end{figure*}

\subsection {Oracle Information Provided  Analysis}

One natural question is how much improvement will be made when the column and table mentioned information (oracle schema information) is provided, As shown in Table \ref{tab:gold}, we report the exact match accuracy of our {\modelname} model on Spider and Spider-SYN benchmarks with oracle schema information provided.  Specifically, we directly change the implicit graph matrix $\widetilde{\mathbf{A}}^{(t)}$ in Eq. \ref{eq: combine} based on the oracle information.  

From Table \ref{tab:gold}, we could discover that: 1) Oracle schema linking information could bring a huge improvement (7.7\% and 17.9\% in Spider and Spider-SYN benchmarks respectively) to the final Text-to-SQL accuracy, which proves the importance of schema linking component. 2) When oracle schema linking information is provided, the results on the Spider-SYN benchmark (80.1\%) will be close to those on the Spider(83.1\%) , which demonstrates that the original performance drops in the Spider-SYN benchmark mainly come from the corrupted schema linking graph. 3) Only providing oracle schema information (which schema item appears) could achieve a similar improvement compared to the oracle schema linking, which could demonstrate our assumption that schema information is more important in the schema linking graph. 4) Oracle column information contributes more than the oracle table information, which is because the column information is more difficult to capture. 

Based on the above results, we can conclude that there is a huge potential for improvement by designing a better schema linking component, which could be an important direction for future work.

\begin{table}[bt!]

\resizebox{0.95\linewidth}{!}{
\begin{tabular}{l|c|c}
\toprule
\multicolumn{1}{c}{\textbf{Model}} & \multicolumn{1}{c}{\textbf{Spider}} & \multicolumn{1}{c}{\textbf{Spider-SYN}}   \\
\midrule 
\midrule 
ISESL-SQL  & 75.8  & 62.2 \\
ISESL-SQL + Oracle columns  & 80.8 (5.0$\uparrow$) & 75.6(13.4$\uparrow$)\\
ISESL-SQL + Oracle tables  &79.1 (3.3$\uparrow$)  &74.3(12.1$\uparrow$) \\
ISESL-SQL + Oracle schema &83.2 (7.4$\uparrow$) &79.5 (17.3$\uparrow$)\\
ISESL-SQL + Oracle schema linking   & \textbf{83.5} (7.7$\uparrow$) &\textbf{80.1}(17.9$\uparrow$)\\
\bottomrule
\end{tabular}}
\caption{Exact match accuracy (\%) on Spider and Spider-Syn benchmarks when oracle schema information is provided.}
\label{tab:gold}
\vspace{-0.3in}
\end{table}

\renewcommand{\arraystretch}{1.3}
\begin{table}[t!]
\centering
\resizebox{1.00\linewidth}{!}{
\begin{tabular}{l}
\toprule
\begin{tabular}{l@{}@{}} 
\multicolumn{1}{@{}p{1.06\linewidth}}{\textbf{Question:} Find the distinct breed type and size type combinations for \textbf{{\color{blue}puppies.}}}
\\
\multicolumn{1}{@{}p{1.06\linewidth}}{\textbf{LGESQL:} \texttt{SELECT breed\_name, size\_code FROM \textbf{{\color{red} Breeds}}}}
\\ 
\multicolumn{1}{@{}p{1.06\linewidth}}{\textbf{ISESL-SQL:} \texttt{SELECT breed\_name, size\_code FROM \textbf{{\color{blue} Dogs}}}}
\end{tabular}     
\\ 
\hline
\begin{tabular}[c]{@{}l@{}}
\multicolumn{1}{@{}p{1.06\linewidth}}{\textbf{Question:} Please show the \textbf{{\color{blue}record type}} of \textbf{{\color{blue}ensembles}} in ascending order of count.}
\\
\multicolumn{1}{@{}p{1.06\linewidth}}{\textbf{LGESQL:} \texttt{SELECT \textbf{{\color{red}performance.Type}} FROM \textbf{{\color{red}performance}} GROUP BY \textbf{{\color{red}performance.Type}} ORDER BY COUNT(*) ASC}} 
\\
\multicolumn{1}{@{}p{1.06\linewidth}}{\textbf{ISESL-SQL:} \texttt{SELECT \textbf{{\color{blue}Major\_Record\_Format}} FROM \textbf{{\color{blue}orchestra}} GROUP BY \textbf{{\color{blue}Major\_Record\_Format}}  ORDER BY COUNT(*) ASC}} 
\end{tabular}    
\\
\hline
\begin{tabular}[c]{@{}l@{}}
\multicolumn{1}{@{}p{1.06\linewidth}}{\textbf{Question:} What is the average and highest \textbf{{\color{blue}capacities}} for all stations?}
\\
\multicolumn{1}{@{}p{1.1\linewidth}}{\textbf{LGESQL:} \texttt{SELECT AVG(stadium.Capacity), \textbf{{\color{red}MAX(stadium.Highest)}} FROM stadium}} 
\\
\multicolumn{1}{@{}p{1.1\linewidth}}{\textbf{ISESL-SQL:} \texttt{SELECT avg(capacity) ,  \textbf{{\color{blue}max(capacity)}} FROM stadium}} 
\end{tabular}    
\\
\hline
\begin{tabular}[c]{@{}l@{}}
\multicolumn{1}{@{}p{1.06\linewidth}}{\textbf{Question:} Of all the competitors who got voted, what is the competitor number and name of the competitor who got \textbf{{\color{blue} least votes}} ?}
\\
\multicolumn{1}{@{}p{1.1\linewidth}}{\textbf{ISESL-SQL:} \texttt{SELECT  contestant\_name FROM contestants JOIN votes ORDER BY \textbf{{\color{red} votes.vote\_id}}  LIMIT 1}} 
\\
\multicolumn{1}{@{}p{1.1\linewidth}}{\textbf{Gold:} \texttt{SELECT contestant\_name FROM contestant JOIN votes \textbf{{\color{blue} GROUP BY contestant\_number ORDER BY  count(*)}}   LIMIT 1}} 
\end{tabular}    
\\
\bottomrule
\end{tabular}
}
\caption{Case study: the first two cases are sampled from Spider-SYN and the last two cases are sampled from Spider-DK. The first three cases are positive cases while the last one is a negative case.
}\label{tab:example}
\vspace{-0.3in}
\end{table}

\subsection {Case study}

To intuitively show the effectiveness of our model, we select four cases from Spider-SYN and Spider-DK benchmarks to compare the generated SQL statements of our {\modelname} model and LGESQL.  The first two cases are from the Spider-SYN benchmark and the last two cases are from the Spider-DK benchmark.  As shown in Table \ref{tab:example}, we can observe that our model could generate correct SQL even in synonym substitution scenario. As in the first case, when the schema-related token \textit{dogs} is replaced with \textit{puppies} in the question, LGESQL fails to identify table name \textit{dogs} while our {\modelname} model could successfully recognize the correct table.  In the third case, the token \textit{highest} in question could be matched to the column \textit{highest} via the exact match based method, which leads to the error in LGESQL. Since our method uses a semantic-based approach to do matching, the above problem could be avoided in most cases. However in the more complicated cases (e.g. the forth case in Table \ref{tab:example}), {\modelname} may fail to recognize the complex structure information.

Figure \ref{fig:case} shows the alignment matrix between question and schema during the training process. We take the sentence \textit{Show the ages for all French Musicians} and the schema in the database \textit{concert\_singer} as an example. The first subfigure shows the alignment matrix generated by the initial graph probing component before training. Although the linking of column `age' and table `singer' is identified, the initial graph probing component still fails to capture the linking between \textit{French} and {country}.  As we can see, during the training process, the graph is iteratively optimized. The correct linking is emphasized in epoch 50. Finally, in epoch 100, our method could almost exclusively focus on the correct linking. The whole process demonstrates that our {\modelname} method could generate better schema linking information iteratively during training.

\section{Related Work}

\subsection{Schema Linking in Text-to-SQL Parsing}
Text-to-SQL Parsing is an essential sub-field of Natural Language Processing and could benefit many other tasks such as Question Answering and Information Extraction  \cite{chen-etal-2017-reading,lin2022inferring,hu2021semi,liu2022hierarchical,hu2021gradient,li2022pair,hu-etal-2020-selfore}.
Schema linking is an indispensable module in recent Text-to-SQL models, which establishes a link between question tokens and schema items \cite{gan-etal-2020-review}.   Many previous works treat schema linking as a minor pre-processing procedure \cite{guo-etal-2019-towards, wang-etal-2020-rat, xu-etal-2021-optimizing, cao-etal-2021-lgesql,li2022multi}, which use surface form match based method to find the occurrences of the schema names in the question. However, these methods may fail in synonym and typo scenarios. Other works implement schema-linking by learning a similarity score between a word and a schema item \cite{bogin-etal-2019-representing, krishnamurthy-etal-2017-neural}, but still suffer from the limitation of the static word embedding. \citet{guo-etal-2019-towards} and  \citet{wang-etal-2020-rat} conduct an ablation study on schema linking, and the results show that removing the extra schema linking causes the biggest performance decline compared to removing other
removable modules. To better investigate the influence of schema linking, \citet{lei-etal-2020-examining} and \citet{taniguchi2021investigation} invest human resources to annotate schema linking information in dataset and employ the full supervision to train Text-to-SQL models.  The result shows that with gold schema linking information as training data, current Text-to-SQL models can exceed the baseline by a very large gap. \citet{dong-etal-2019-data} utilize implicit linking supervision to train their schema linking model with reinforcement learning method. 
\citet{liu-etal-2021-awakening} train the schema linking module using pseudo alignment as supervision from PLMs, but the pseudo alignment could still be noisy. Our {\modelname} iteratively refines the schema linking graph using both supervision from SQL generation task and the schema mention information in SQL statements.
\subsection{Graph Learning based Models}
Graph neural networks (GNNs) are neural models that capture the dependence of graphs\cite{zhou2020graph}. The structure of the graph is normally labeled by experts \cite{sen2008collective} , pre-processed with existing relation parsing tools \cite{wu2021graph} or precomputed by the k-nearest neighbor algorithm \cite{anastasiu2015l2knng}. However, the graph structure generated by these methods may contain missing or noisy edges which may not be optimal for downstream learning tasks.  To alleviate this problem, 
some works propose a robust graph learning schema by removing noise and errors in the raw data adaptively \cite{kang2019robust}.  Similar to these works, 
graph attention network is proposed to use self-attention mechanism to reweight edge importance \cite{DBLP:conf/iclr/VelickovicCCRLB18}.  As these robust learning approaches still cannot handle the missing edges,
adaptive graph structure learning methods are proposed to facilitate downstream graph-based tasks \cite{jiang2019semi}. These adaptive graph construction approaches typically utilize a parameterized graph similarity metric learning function to learn an adjacency matrix by considering pair-wise node similarity in the embedding space \cite{wu2021graph}. These models only perform graph structure learning for one time which is not enough. To better optimize graph structure and downstream tasks jointly, 
 \citet{chen2020iterative} proposed an iterative deep graph learning model to let graph learning models and downstream tasks optimize together.  In the scenarios when the input graph is not available,  LDS \cite{franceschi2019learning} is proposed to learn the graph structure by solving a bilevel program that learns the discrete probability over the graph egdes.  However, these models cannot be applied to Text-to-SQL directly because the graph in Text-to-SQL model is heterogeneous which contains different types of nodes and edges.  In this paper, we propose a framework to introduce the graph structure learning network into a modified version of Relation-aware graph attention network  \cite{wang-etal-2020-relational}  to enhance Text-to-SQL models.

\section{Conclusion}



In this paper, we propose a framework named {\modelname} to build a semantic enhanced schema-linking graph for the Text-to-SQL task. Different from the previous exact match-based schema linking method, the schema linking graph generated by our method is  accurate and robust in various settings such as synonym substitution. We perform extensive experiments on three benchmarks and the results demonstrate the effectiveness of our method.

Here, we list several main findings as follows. 1) The schema linking graph generated by PLM could help the Text-to-SQL model find the correct schema information.  2) Implicit graph learning module plays a very important role to capture the correct schema linking information. 3) In the schema linking graph, the correctness of the schema items is much more important than the question tokens, which indicates that weak supervision could work in most scenarios. 4) When oracle schema information is provided, the Text-to-SQL accuracy will be greatly improved, which implies a huge potential for  improvement. 5) The auxiliary task graph regularization could consistently yield improvements on multiple benchmarks.


\section*{ACKNOWLEDGMENTS}

The work was supported by the National Key Research and Development Program of China (No. 2019YFB1704003), the National Nature Science Foundation of China (No. 62021002 and No. 71690231), Tsinghua BNRist and Beijing Key Laboratory of Industrial Big Data System and Application.

\bibliographystyle{ACM-Reference-Format}
\bibliography{sample-base}
\appendix

\section{Decoder architecture}

Given the question representation $(\mathbf{q}_{1}^{e}, \mathbf{q}_{2}^{e}, \ldots, \mathbf{q}_{|Q|}^{e})$ and schema representation $(\mathbf{s}_{1}^{e}, \mathbf{s}_{2}^{e}, \ldots, \mathbf{s}_{|T|+|C|}^{e})$ generated by graph encoder module, Text-to-SQL decoder module aims to generate the corresponding SQL statements.

The architecture of our Text-to-SQL decoder is similar to previous work \cite{yin-neubig-2017-syntactic}, which generates the abstract syntax tree (AST) of the target SQL $y$ in depth-first traversal order.  The decoder utilizes a LSTM to generate actions in each timestep which is either: (1) An APPLYRULE action that expands the current non-terminal node based on SQL grammar in the partially generated AST. (2) A SELECTTABLE or SELECTCOLUMN action that chooses one table or column item from the encoded schema representation $(\mathbf{s}_{1}^{e}, \mathbf{s}_{2}^{e}, \ldots, \mathbf{s}_{|T|+|C|}^{e})$. 

Formally, the whole process can be defined as
\begin{equation}
P(y \mid \mathbf{X})=\prod_{i} P\left(a_{i} \mid a_{<i}, \mathbf{X}\right)
\end{equation}
where $X$ is the combined question and schema representation,  $a_{<i}$ is all the previous actions before step i.
Specifically, in each step, the decoder network updates the state as follows:
\begin{equation}
\mathbf{m}_{i}, \mathbf{h}_{i}= LSTM \left(\left[\mathbf{a}_{i-1} ; \mathbf{a}_{p_{i}} ; \mathbf{h}_{p_{i}} ; \mathbf{t}_{i}\right], \mathbf{m}_{i-1}, \mathbf{h}_{i-1}\right)
\end{equation}
where $\mathbf{m}_{i}$ and $\mathbf{h}_{i}$ are the cell state and output of the i-th timestep, $\mathbf{a}_{i-1}$ is previous action embedding, $\mathbf{a}_{p_{i}}$ is the parent embedding of current node, $\mathbf{h}_{p_{i}}$ is the parent cell state, $\mathbf{t}_{i}$ is the type embedding of current node.  

To this end, the operation APPLYRULE can be represented as 
\begin{equation}
P\left(a_{i}=\text { APPLYRULE }[R] \mid a_{<i}, \mathbf{X}\right)= 
\operatorname{softmax}_{R}\left(\mathbf{h}_{i} \mathbf{W}_{\mathrm{R}}\right)
\end{equation}
where $\mathbf{W}_{\mathrm{R}}$ transforms the LSTM output into action rule logits.  

The SELECTTABLE is implemented by calculating the attention between LSTM hidden state $\mathbf{h}_{i}$ and table representation $\mathbf{x}_{t_{j}}$:
\begin{equation}
\begin{split}
P\left(a_{i}=\operatorname{SELECTTABLE}\left[t_{j}\right] \mid a_{<i}, \mathbf{X}\right)= \\
\operatorname{softmax}_{j}\left(\mathbf{h}_{i} \mathbf{W}_{tq}\right)\left(\mathbf{x}_{t_{j}} \mathbf{W}_{t k}\right)^{\mathrm{T}}
\end{split}
\end{equation}
And the SELECTCOLUMN operation is defined in a similar way.

The loss of Text-to-SQL is defined as follows:
\begin{equation}
\mathcal{L}_{\mathrm{SQL}}= \sum_{i=1}^{T} \log P\left(a_{i} \mid a_{<i}, \mathbf{X}\right)
\end{equation}
where $T$ is the total number of actions in the abstract syntax tree.

\end{document}